# Multi Objective Particle Swarm Optimization based Cooperative Agents with Automated Negotiation

Najwa Kouka, Raja Fdhila and Adel M. Alimi

REGIM-Lab.: REsearch Groups in Intelligent Machines, University of Sfax, National Engineering School of Sfax (ENIS), BP 1173, Sfax, 3038, Tunisia
{najwa.kouka.tn, raja.fdhila, adel.alimi}@ieee.org

**Abstract.** This paper investigates a new hybridization of multi-objective particle swarm optimization (MOPSO) and cooperative agents (MOPSO-CA) to handle the problem of stagnation encounters in MOPSO, which leads solutions to trap in local optima. The proposed approach involves a new distribution strategy based on the idea of having a set of a sub-population, each of which is processed by one agent. The number of the sub-population and agents are adjusted dynamically through the Pareto ranking. This method allocates a dynamic number of sub-population as required to improve diversity in the search space. Additionally, agents are used for better management for the exploitation within a sub-population, and for exploration among sub-populations. Furthermore, we investigate the automated negotiation within agents in order to share the best knowledge. To validate our approach, several benchmarks are performed. The results show that the introduced variant ensures the trade-off between the exploitation and exploration with respect to the comparative algorithms.

**Keywords:** Multi Objective Optimization Problems, Particle Swarm Optimization, Multi Agent System, Distributed Architecture, Automated Negotiation

## 1 Introduction

Optimization problems have received appreciable attention over the past decade and presented as an active research field that is encountered in various fields of technology such as image processing [1], path planning [2] and handwriting recognition [3-8]. The optimization can be incorporated into other intelligent tools of soft computing such as the neural network [9,10] and the fuzzy system [11] to produce better and faster result. In fact, Swarm intelligence (SI) is considered as an adaptable concept for the optimization problem. One of the most dominant algorithms in SI is a particle swarm optimization (PSO). Although PSO has been widely used for solving many well-known numerical test problems, but it suffers from the premature convergence. Thereby, several strategies have been developed responding to this limitation, such as the distributed evolutionary (DE) [12,13]. Hence, with the DE, a parallel optimization process can be formed which offers the ability to resolve a high-dimensional problem. Frequency, the DE presented at the population level, in which the population is distributed within the search space. The DE increase the diversity of solutions, thereby solve the premature convergence. Due to the several issues that address the distributed sub-populations, such as the communications protocol, it becomes required to endow a novel system with the capability to communicate, cooperate and reach agreements within the different sub-populations. These trends have led to the incorporation of Multi-Agent System (MAS) [14] as a distributed model. MAS allows building a distributed PSO with greater ease and reliability. In this work, we investigate a new distributed MOPSO based cooperative agents (MOPSO-CA) to optimize MOP. The MAS is advantageously used to elaborate a new variant of distributed MOPSO. Additionally, applying the automated negotiation [15], in order to share the best knowledge among sub-populations. In this way, the good information obtained by each sub-population is exchanged among the sub-populations; thereby the diversity of the population is increased simultaneously.

The organization of the remainder of this paper is as follows: Then, we introduce the main concepts of our approach. We present the related work in section 3. Section 4 details the purpose of our approach. The experimental result is discussed in section 5. Finally, the conclusion and future work are then summarized in section 6.

## 2     Theoretical Foundation

In this section, we briefly present the main concept that will be employed throughout this article. First, we need to define a multi-objective optimization problem (MOP) and its basic concept, then the MOPSO.

### 2.1     Introduction to Multi Objective Optimization Problem

A MOP has a number of objective functions, which are to be minimized or maximized simultaneously. Those objectives are often immeasurable and conflicting with each other. MOP typically contains a set of constraints, that any feasible solution must satisfy, including the set of the optimal solution [16]. Subsequently, MOP can be written mathematically as follow:

$$\begin{cases} min/\max(f_i) & i = 1,2,\dots,k \\ g_j(x) \geq 0 & j = 1,2,\dots,J \\ h_p(x) \geq 0 & p = 1,2,\dots,H \\ x_i^l \leq x_i \leq x_i^u & i = 1,2,\dots,n \end{cases} \quad (1)$$

In order to define the concept of optimization, we introduce a few useful terminologies:
**Dominance relationship.** A solution x dominates solution y, if x is no worse than y in all objectives, and x is strictly better than y in at least one objective.
**Pareto optimal.** Is a non-dominated solution, which are equally good when compared to other solutions, means there exists no other feasible solution, which would decrease some criterion without causing a simultaneous increase in at least one other criterion.
**Pareto front.** The plot of the objective functions whose non-dominated vectors are in the Pareto optimal set is called the Pareto.

**Convergence.** The Pareto-front, which is as close as to the true Pareto-front, is considered best. Ideally, the true Pareto-front should contain the best-known Pareto-front.
**Diversity.** Pareto-front should provide solutions, which are uniformly distributed and diversified across the Pareto-front.

## 2.2  Introduction to Multi Objective Particle Swarm Optimization

PSO is a population-based search algorithm introduced by Kennedy and Eberhart [17]. In PSO, each particle in the population is a solution to the problem. Instead, the particles are "flown" through hyper dimensional search space to search out a new optimal solution through two equations (2) and (3).

$$\vec{v}_i(t) = W\,\vec{v}_i(t-1) + C_1 r_1\left(\vec{x}_{pbest_i} - \vec{x}_i(t)\right) + C_2 r_2(\vec{x}_{leader} - \vec{x}_i) \qquad (2)$$

$$\vec{x}_i(t) = \vec{x}_i(t-1) + \vec{v}_i(t) \qquad (3)$$

Where leader presents the global best solution in the population, the pbest position is the best personal solution of a given particle. Additionally, $r_1$ and $r_2$ are random values, W is the inertia weight, $c_1$ is the cognitive learning factor and $c_2$ is the social learning factor. In order to adopt the PSO for the optimization of MOP (MOPSO), a few modifications must be made under the original PSO. First, the aim is to discover a set of the optimal solution, not even one. Second, an external archive is kept, where all non-dominated solutions found at each iteration are saved in.

## 3  Related Works

MOPSO is one of the dominant techniques to find the promised solutions much faster than the other algorithms. Instead, it suffers from the premature convergence. This problem tends to converge to local optima, such that problem led the MOPSO to fail to find the Pareto-optimal solutions. In order to solve this problem, it is obvious that the original algorithm has to be modified. One of the interesting methods that have the ability to overcome the problem of premature convergence is the distributed evolutionary (DE). The granularity of the DE may be at the population level. Since the distributed population based on the idea of dividing the entire population into sub-populations, each of which is processed by one processor. In fact, we summarize a few outputs.
A new version of MOPSO [18] adopts the Pareto ranking to dynamic subdivide the population. There are a few variants of hierarchical architecture are proposed in [19-21] that is proposed by Fdhila. Its main idea is to have a 2-levels that adopts a bidirectional dynamic exchange of particles between MOPSOs. Indeed, these variants improve its efficacy in many real applications such as the feature selection [22], the routing Pico-satellites problem [23], the grasp planning problem [24], the TSP problem [2], the Face Recognition [25]. The organization and communication between sub-population play an important role in the DE. In fact, there are varieties of methods that attempt to address this deficiency. One of the most notable methods is the incorporation of MAS as a model of DE. Indeed, the MAS is adopted to model, manage and coordinate the process of optimization among different sub-population. Different methods are improved in [26-29] these methods applied the MAS as a model of DE, in which the MAS achieves the purpose of communication, organization and cooperation.

## 4       Description of the Proposed Approach

### 4.1     Motivation

The adaptation of MOPSO with DE makes evident the notion of using MAS could be the straightforward way to recover MOPSO in order to overcome the premature convergence. The hybridization between the MAS and the MOPSO algorithm could balance between the local exploitation and the global exploration. Indeed, we propose to use not one, but several sub-populations (each with a dynamic size). Each sub-population overfly within a specific region of the search space. In addition, it has its own set of particle and particle guides kept in the local archive. It is known that the use of disconnected sub-population led the algorithm unable to converge to the true Pareto front. This issue makes the using of a good strategy of communication is necessary. In this context, the automated negotiation (AN) is used. The AN used to ensure the changing information, considering that we are instead a decision conflict between agents, which are the optimal solution to be selected. In fact, the AN accurate the selection of global leader, since the solution is largely depends on the guiding points. In this way, the AN guarantees the set of Pareto optimal solution since each agent tends to exploit the sub-search space, while ensuring that an exploration is reached within the search space by sharing best knowledge (among sub-populations).

### 4.2     Main Process

The main process of our algorithm is illustrated in Figure 1. At the first stage, the population creates, initializes and updates its own particles. Additionally, the leaders set is generated and saved in the external archive. Once the initialization is completed, the Pareto ranking divides the population, as a result, a dynamic number of fronts (F0, F1..., Fn) are generated. Each of these fronts plays the role of the sub-population. These sub-populations distributed among agents. Then, for a maximum number of iterations, each agent performs the execution of MOPSO in its own sub-population, including the selection of gbest (global leader) (see Fig. 2) by the AN, the update of position and velocity and, finally, the local archive is updated too. However, as AN process we adopt a multi-lateral negotiation since we have k cooperative agents as negotiators. In this model, we assume that our domain has one issue (defining the best solution in order to share among sub-populations).

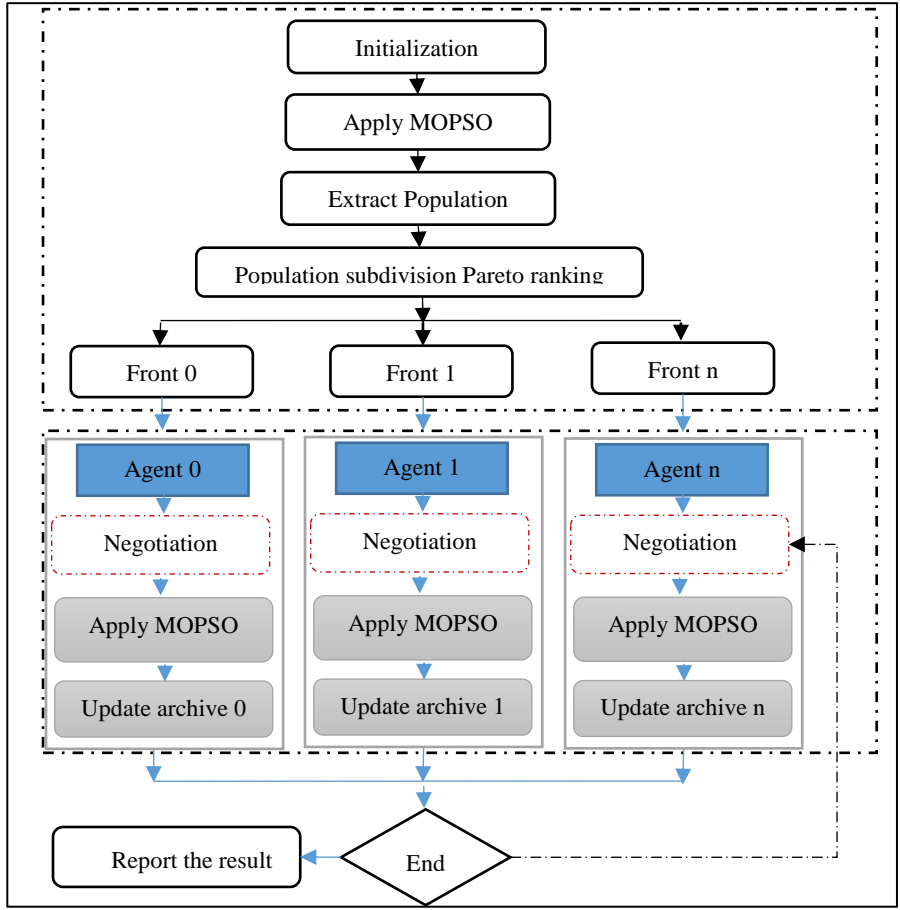

**Fig. 1.** Flowchart of MOPSO-CA algorithm

The general process of the negotiation process is as follows: once all agents attending the evolving process, a new negotiation session has begun, first, each agent sends a call for proposal CFP to other agents. Next, the agent responds to the CFP by making an offer (best local solution: based on the dominance operator). In the turn, the agent evaluates the incoming offer using the fitness function. Consequently, the accepted proposal is the offer which accepted by all agents. In result, the accepted proposal becomes the best global solution that used during the update position. Therefore, each particle of sub-population adjusts its trajectory according to its own experience ($p_{best}$), the experience of its neighbors ($l_{best}$), and the experience of best global solution among sub-populations ($g_{best}$). So the new equation for velocity is presented in equation (4).

$$\vec{v_i}(t) = w\,\vec{v_i}(t-1) + c_1 r_1 \left(\vec{x}_{pbest_i} - \vec{x}_i(t)\right) + c_2 r_2 \left(\vec{x}_{lbest} - \vec{x}_i\right) + c_3 r_3 \left(\vec{x}_{gbest} - \vec{x}_i\right) \quad (4)$$

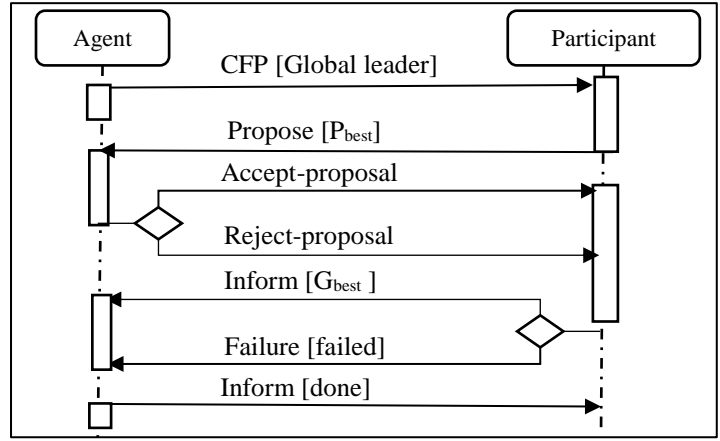

**Fig. 2.** Negotiation protocol

## 5   Experimental studies

In order to know how important MOPSO-CA was, we compared it against two algorithms that taken from the literature [30] of MOP's algorithms namely: Non-dominated Sorting Genetic Algorithm II (NSGAII), and Optimized Multiobjective Particle Swarm Optimization (OMOPSO). The benchmarks were explored in our experiment are DTLZ family (DTLZ5 and DTLZ6) and UF family (UF1, UF2, UF3 and UF10), which have sufficient complexity to evaluate the algorithm's performance, in terms of solution diversity and convergence rate.

### 5.1   Performance Metric

Several performance evaluations are available to compare the performance of the presented approach. In the present context, we choose the following three metrics [31]: Spread (SP), Inverted Generational Distance (IGD) and Hypervolume (HV) which used to evaluate the diversity, the convergence and the both (convergence and diversity) respectively.

### 5.2   Experimental setting

To evaluate the performance of the comparative algorithms, 30 runs of each algorithm for each test function are performed; a population with 200 individuals is fixed, and the archive size is set to 100. Further, the parameters of different algorithms detailed as the following, for MOPSO-CA and MOPSO, an acceleration coefficients $c1, c2$ = Rand (1.5, 2.0), and inertia weight $w$ = Rand (0.1, 0.5). For NSGAII the max evaluations = 25000 and crossover probability = 0.9.

### 5.3 Experimental results

In this section, we analyze the results obtained by the algorithms. Derivatives figures (Fig.3 to Fig.6) show the graphical results generated by the comparative algorithms. (see Fig.3 and Fig.4) show the Pareto front produced by the comparative algorithms for DTLZ5 and UF10 respectively; clearly, we can conclude that NSGAII, OMOPSO and MOPSO-CA cover the entire true Pareto front of DTLZ5 on one hand. On the other hand, we can see that only the MOPSO-CA may cover the true Pareto front of UF10.

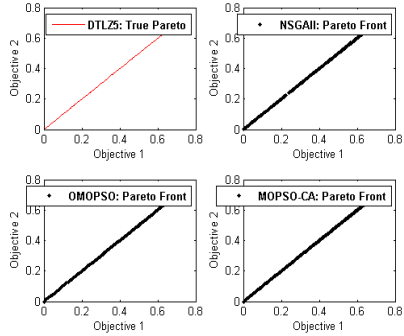
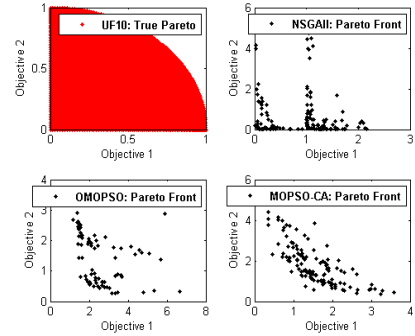

**Fig. 3.** Pareto fronts obtained by the algorithms for DTLZ5

**Fig. 4.** Pareto fronts obtained by the algorithms for UF10

For more precise the accuracy of the solution, statistical values are provided in Table 1. From these values, it can be seen, that the average performance of MOPSO-CA is better than NSGAII and OMOPSO with respect to the HV metric.

**Table 1.** Performance metrics (Mean value) for the different test functions

|       | Metric | NSGAII | OMOPSO | MOPSO-CA |
|-------|--------|--------|--------|----------|
| **UF1** | SP  | 1:37  | 9:77e | **1:32e** |
|       | IGD  | 1:85e | 1:99e | **0:00e** |
|       | HV   | 6:37e | 6:05e | **9:53e** |
| **UF2** | SP  | 7:77e | **5:18e** | 6:40e |
|       | IGD  | 1:28e | **1:11e** | 3:79e |
|       | HV   | 7:02e | 7:04e | **7:25e** |
| **UF3** | SP  | 9:82e | 6:62e | **6:56e** |
|       | IGD  | 1:09e | 6:43e | **0:00e** |
|       | HV   | 2:43e | 3:85e | **4:31e** |
| **UF10** | SP | 8:63e | 6:79e | **6:20e** |
|       | IGD  | 7:57e | 2:18e | **0:00e** |
|       | HV   | 6:14e | 1:64e | **7:14e** |
| **DTLZ5** | SP | 4:57e | 1:82e | **1:68e** |
|       | IGD  | 5:78e | 8:15e | **3:95e** |
|       | HV   | 9:40e | 9:32e | **9:57e** |
| **DTLZ6** | SP | 7:98e | 1:28e | **1:03e** |
| | IGD | 2:69e | 7:41e | **2:13e** |
| | HV | 0:00e | 9:35e | **9:48e** |

Regarding the SP metric, we can conclude that MOPSO-CA has the better spread of solutions for UF1, UF3, UF10, DTLZ5 and DTLZ6. On the other hand, the OMOPSO has the best SP value for UF2. Regarding the IGD metric, we can conclude that MOPSO-CA is relatively better than other algorithms for UF1, UF3, UF10, DTLZ5 and DTLZ6, since MOPSO-CA have the minimum IGD values. On the other hand, OMOPSO was the best for UF2. Hence, to have a deep dissection of the MOPSO-CA, the HV values for UF1 and UF2, DTLZ5 and DTLZ6 were plotted (see Fig.5).

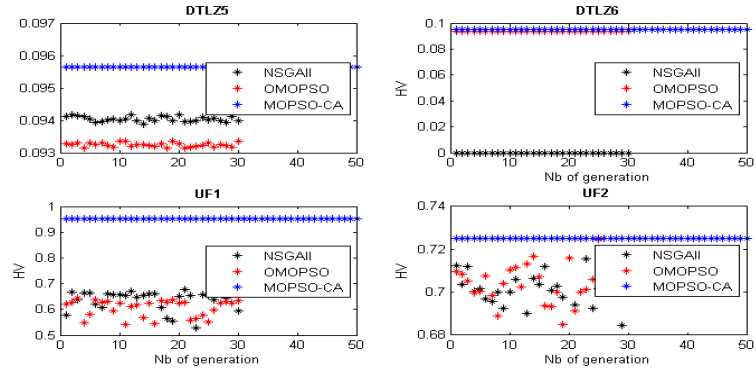

**Fig. 5.** Performance (HV) over DTLZ5, DTLZ6, UF1 and UF2 problem

Graphically, it is intuitive that MOPSO-CA can achieve the best tradeoff between convergence and diversity (the higher value, the better performance for HV) with respect to other algorithms. Meanwhile, according to the HV values, we can conclude that MOPSO-CA gets better performance of different test function. Clearly, our MOPSO-CA produces the best trade-off between the convergence and diversity, within the tested problems.

## 6    Conclusion and Future Work

In this paper, the MOPSO-CA algorithm proposed to solve MOP. In this algorithm, the sub-populations, Pareto ranking, MAS and automated negotiation are used. MAS improve the performance of distributed MOPSO but strategies for communication between agents are very important. Thus, the efficiency of synchronous knowledge (most successful solution) exchange strategies has been achieved by using the automated negotiation. Through experiments, it can be concluded that MOPSO-CA can outstanding performances in terms of convergence and diversity qualities. As a future work, we will explore more the feature of MAS to increase the intelligence level of particles. In addition, our proposed approach can be incorporated in many real-world problems.

**Acknowledgement.** The research leading to these results has received funding from the Ministry of Higher Education and Scientific Research of Tunisia under the grant agreement number LR11ES48.